%% file: main.tex
\documentclass[runningheads]{llncs}
\usepackage[T1]{fontenc}
\usepackage{graphicx,verbatim}

\usepackage{hyperref} 
\usepackage{amsmath}

\usepackage{multirow} 
\usepackage{booktabs}
\usepackage{subcaption}
\usepackage{enumitem} 
\usepackage[capitalize,nameinlink]{cleveref}
\crefname{section}{Sec.}{Secs.}
\Crefname{section}{Section}{Sections}
\Crefname{table}{Table}{Tables}
\crefname{table}{Tab.}{Tabs.}
\usepackage{hyphenat} 
\usepackage{floatrow}
\usepackage{color}

\begin{document}
\title{Discrepancy Minimization Improves Cross-Hospital Robustness in Digital Pathology}
\titlerunning{Discrepancy Minimization Improves Robustness in Digital Pathology}
%
\author{Ben Vardi\inst{1} \and
Dana Schonberger\inst{2} \and
Yuval Friedmann\inst{2} \and
Zohar Yakhini\inst{1,3} \and \\
Iris Barshack\inst{2} \and
Alexander Loebel\inst{2}\textsuperscript{*} \and
Ariel Shamir\inst{1}\textsuperscript{*}}
\authorrunning{Vardi et al.}
%
\institute{Reichman University, Herzliya, Israel \and
Institute of Pathology, Sheba Medical Center, Ramat-Gan, Israel \and 
Technion - Israel Institute of Technology, Haifa, Israel\\
\textsuperscript{*} Equal advising contribution
}
  
\maketitle

\input{sec/0_abstract}

\input{sec/1_intro}
\input{sec/2_related_work}
\input{sec/3_method}

\input{sec/4_experiments}
\input{sec/5_conclusions}

\bibliographystyle{splncs04}
\bibliography{bib}

\end{document}

%% file: sec/0_abstract.tex
\begin{abstract}

Pathology foundation models (PFMs) have advanced rapidly in recent years and support training classifiers for a range of histopathology tasks. However, their robustness across hospitals remains limited: performance often degrades when training a classifier on data from one hospital and evaluating it on another target hospital. We address this challenge by fine-tuning PFMs with a local maximum mean discrepancy (LMMD) objective that applies to two settings: domain adaptation, where unlabeled target-hospital data is available, and domain generalization, where target-hospital data is unavailable at all. Experiments at both the patch- and slide-level show consistent improvements across multiple PFMs and tasks.

\keywords{Cross-hospital robustness \and Digital pathology}

\end{abstract}

%% file: sec/1_intro.tex
\section{Introduction}

Digital pathology has witnessed an impressive leap with the emergence of modern pathology foundation models (PFMs)~\cite{chen2024towards,lu2024visual,vorontsov2024foundation}. 
These models can be utilized to process whole-slide images (WSIs) into general-purpose representations for training classifiers on downstream histopathology tasks, such as cancer detection and biomarker prediction,
and may even approach human expert performance~\cite{liu13automated}.

Yet, PFMs may struggle under real-world deployment conditions.
One such problem is that their representations are not robust to variability between hospitals where tissue was collected~\cite{komen2025robustfoundationmodelsdigital}.
For example, performance degrades when classifiers are trained on data from one hospital and evaluated on a different one~\cite{bonn2026cracks,umer2026multicenterbenchmarkmultipleinstance}.
However, labeled data of a specific hospital is often unavailable due to the cost and time required for expert annotation, while labeled data from other hospitals (e.g., from public datasets) may be available.
In the following, we refer to a hospital of interest as a \emph{target hospital}, and hospitals with available labeled data as \emph{source hospitals}.

While this limitation may be alleviated by retraining~\cite{vray2024distill,asadi2024learning,fuhlert2025systematic} PFMs with objectives and data aimed at reducing such performance degradation, these approaches are often impractical due to high compute and data costs.
In this paper, we present a domain adaptation method that can adapt PFMs trained on source hospitals to also perform well on a target hospital in a lightweight manner, using a small amount of data, 
and without the need for annotating the target samples.

Our work leverages the fact that many PFMs~\cite{chen2024towards,lu2024visual,vorontsov2024foundation} operate as patch-level encoders: they encode image patches extracted from WSIs, while WSI-level predictions are obtained by aggregating patch embeddings using multiple instance learning (MIL) methods~\cite{ilse2018abmil}.
Thus, we propose to apply domain adaptation to PFMs at the patch-level, and obtain improved slide-level classification without explicit slide-level adaptation.
This follows the patch-level training paradigm of these models, and offers a simpler approach than pathology-specific adaptation.

\input{sec/tables_and_figures/figure_demo}

In the unsupervised domain adaptation setting, we assume patch-level labeled data from at least one source hospital and unlabeled data from a target hospital of interest.
To adapt the model to the target hospital, we perform LoRA fine-tuning~\cite{hu2021lora} of the PFM while jointly training a classifier on the source labels.
Fine-tuning uses local maximum mean discrepancy (LMMD)~\cite{zhu2020deep}, a domain adaptation objective that aligns the source and target distributions (see~\cref{fig:da_method}).
We additionally consider a harder domain generalization setting, in which the target hospital is unknown in advance and the goal is to improve generalization to unseen target hospitals.
In this case, training uses labeled source data from at least two hospitals, with LMMD applied to encourage alignment across source domains while jointly learning a classifier (see~\cref{fig:dg_method}).
Lastly, fine-tuned PFMs can be used for slide-level tasks (see~\cref{fig:slide_method}). 
We refer to our application of LMMD in pathology as \emph{PFM-LMMD}.
Importantly, PFM-LMMD is PFM agnostic and lightweight, using parameter-efficient LoRA and 
can operate with limited labeled patch data.

Our experiments in both domain adaptation and domain generalization, across two pathology classification tasks and two PFMs, show that PFM-LMMD consistently improves over original PFMs and prior methods on target domains.
Gains are demonstrated at the patch-level, and carry over to slide-level tasks.

Our contributions can be summarized as follows:
\begin{itemize}[label=\textbullet]
    \item We propose PFM-LMMD, an application of the LMMD objective to improve the cross-hospital robustness of PFMs.
    PFM-LMMD can be applied to two scenarios common in practice: when unlabeled target data is available (domain adaptation) or unavailable in advance (domain generalization).
    \item We demonstrate that PFM-LMMD yields gains on target hospitals across several tasks and PFMs, in both scenarios, and at the patch and slide levels.
\end{itemize}

%% file: sec/tables_and_figures/figure_demo.tex
\begin{figure}[!t]
    \centering

    \begin{subfigure}[b]{0.49\textwidth}
        \centering
            \includegraphics[width=\linewidth]{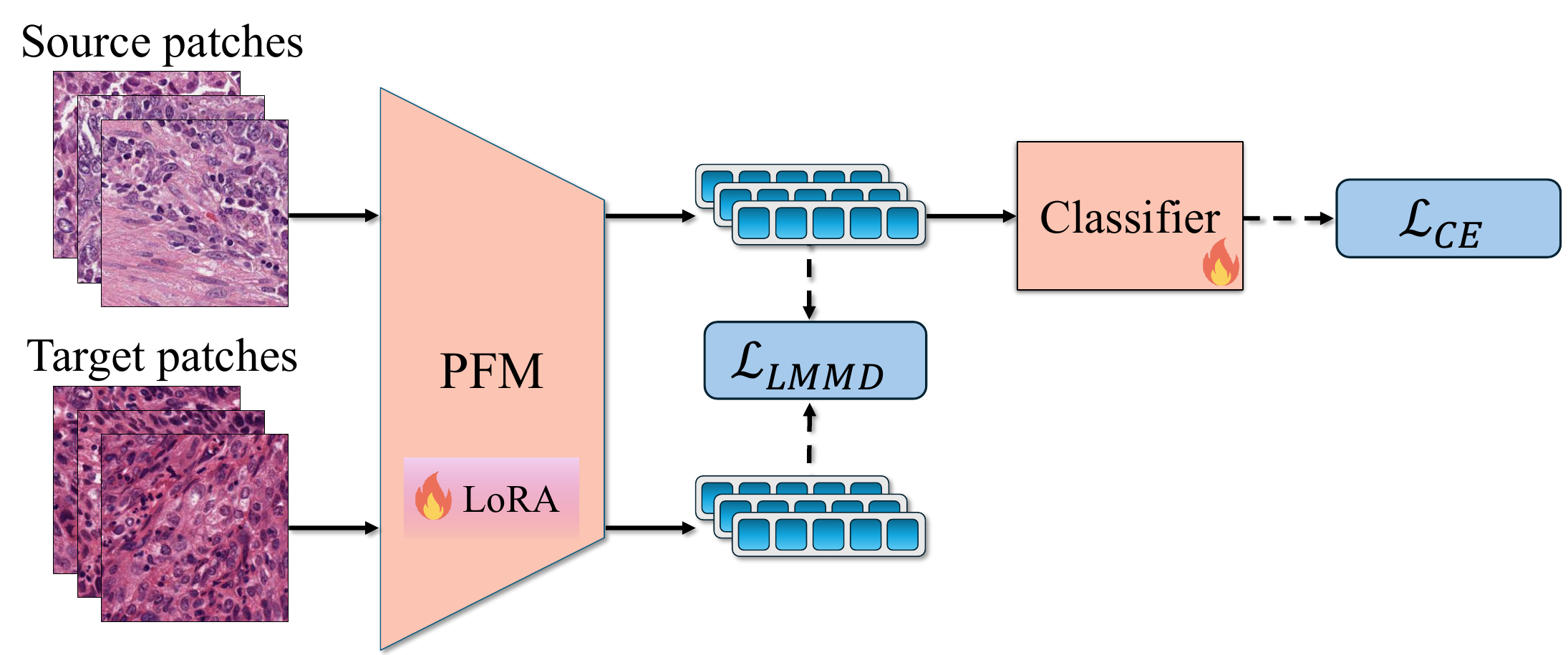}
        \caption{}
        \label{fig:da_method}
    \end{subfigure}
    \hfill
    \begin{subfigure}[b]{0.49\textwidth}
        \centering
                    \includegraphics[width=\linewidth]{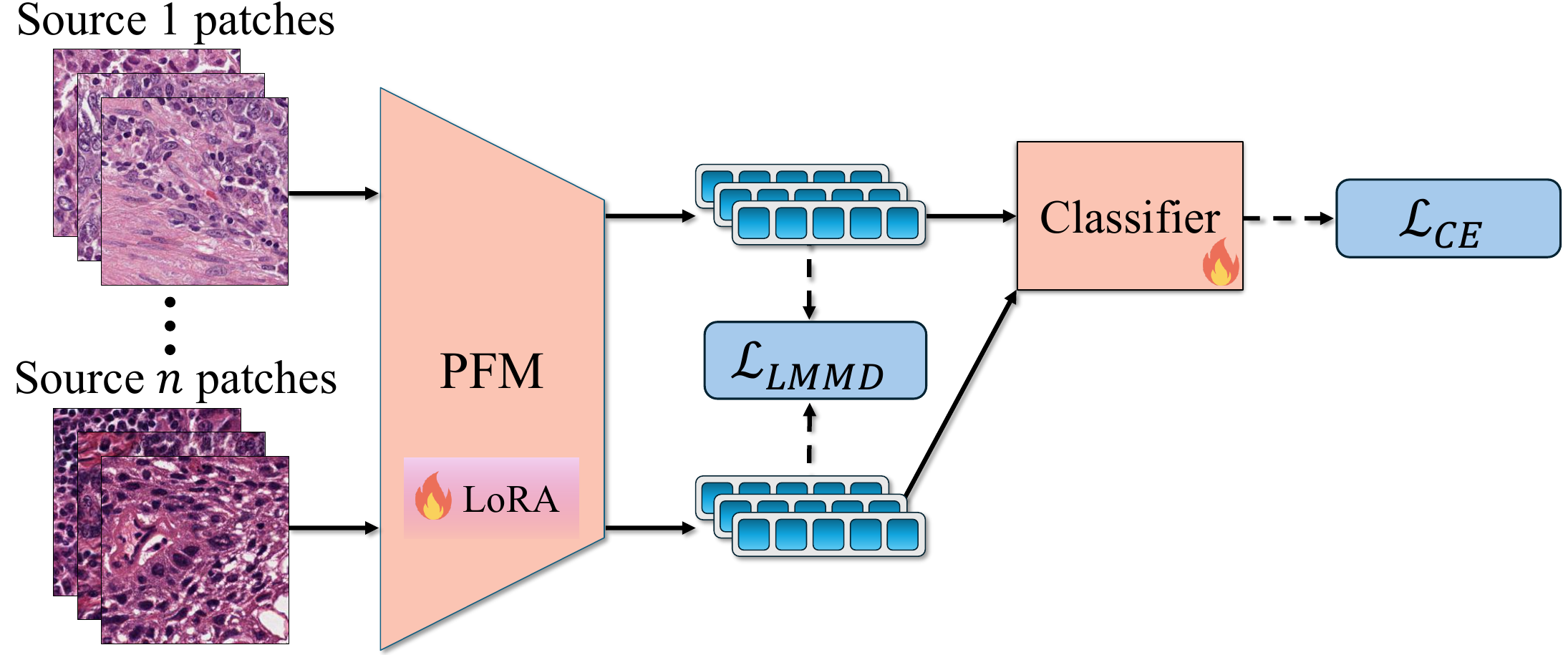}
        \caption{}
        \label{fig:dg_method}
    \end{subfigure}
    \begin{subfigure}[b]{\textwidth}
        \centering
        \includegraphics[width=0.72\linewidth]{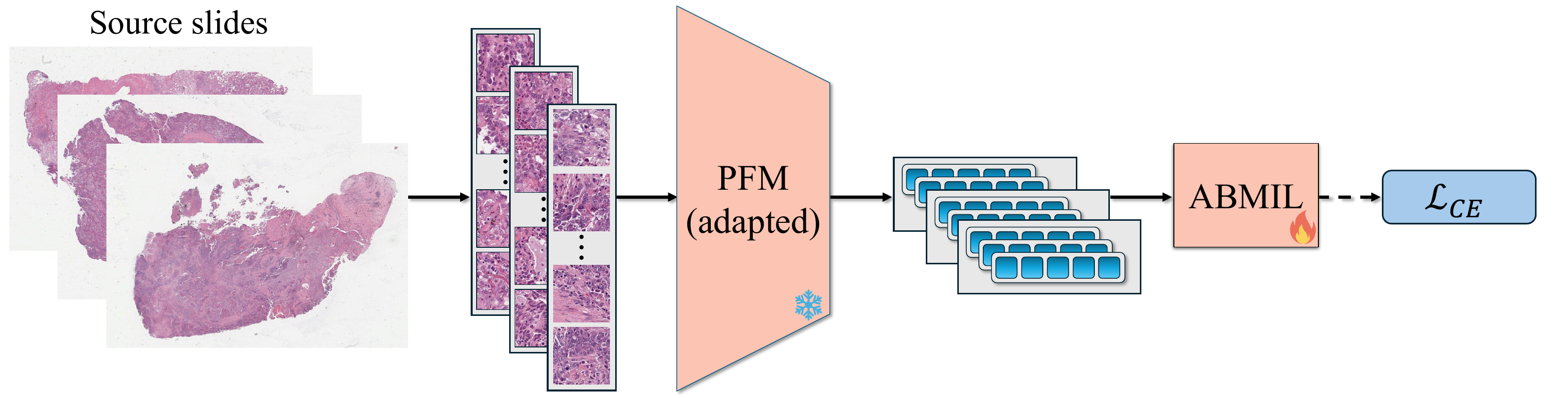}
        \caption{}
        \label{fig:slide_method}
    \end{subfigure}
    \caption{PFM-LMMD training scheme at the patch level for (a) domain adaptation and (b) domain generalization, and (c) at the slide level.}
    \label{fig:method}
\end{figure}

%% file: sec/2_related_work.tex
\section{Related Work}
\label{sec:related_work}

\subsubsection{Pathology Foundation Models Robustness}
\label{subsec:related_work:pfms}

Digital pathology has seen rapid progress in recent years with the rise of transformer\hyp{}based PFMs~\cite{chen2024towards,lu2024visual,vorontsov2024foundation}. These models encode hematoxylin and eosin (H\&E) stained histopathology images into embeddings that can be used for a range of downstream tasks.
Despite their impressive performance, PFMs may face limitations in real-world settings.
A key limitation arises in multi-hospital environments, where H\&E appearance varies due to differences in tissue processing, staining protocols, and slide digitization systems, leading to systematic visual variability across sites~\cite{howard2021impact,thiringer2026scannerinduceddomainshiftsundermine}. 

Recent work has explored various aspects of this robustness problem.
Kömen et al.~\cite{komen2025robustfoundationmodelsdigital} examined hospital-specific signatures in PFM embeddings and performance drops caused by hospital-label bias in classifier training data.
Others~\cite{bonn2026cracks,huang2025knowledge,liu2025hasd} focused on a practical aspect:
learning a classifier for a target hospital when data from that hospital is unlabeled or is unavailable at all.

Various methods have been proposed to address this cross-hospital robustness challenge.
One line of work applies stain normalization~\cite{macenko2009method,reinhard2002color} as test-time preprocessing 
to transform target images to the appearance of the training source data.
Another approach trains pathology models with domain adaptation objectives, either from scratch or from general-purpose pretrained models~\cite{vray2024distill,fuhlert2025systematic}.
While this can be a reasonable choice for task-specific models~\cite{fuhlert2025systematic}, it is less scalable than adapting pretrained PFMs.
Indeed, others proposed to operate on top of frozen PFMs, such as learning a projection head to reduce hospital-specific signatures in embeddings~\cite{komen2025robustfoundationmodelsdigital,carloni2025pathology}, but these were found to provide limited gains.
 
A notable attempt is HASD~\cite{liu2025hasd}, a pathology-specific domain adaptation method that operates at the slide-level, on top of PFM embeddings.
FLEX~\cite{huang2025knowledge} addresses domain generalization by learning an information bottleneck module, but is constrained to vision-language PFMs.
Unlike prior pathology-specific solutions, 
we (1) address the problem with a general-purpose LMMD objective at the patch level, which can be transferred to the slide-level, 
(2) fine-tune the PFM itself rather than operating on top of PFM embeddings, and (3) consider both domain adaptation and generalization rather than a single setting.

\subsubsection{Domain Adaptation and Generalization}
\label{subsec:related_work:da_dg}

The machine learning literature has extensively studied domain adaptation and domain generalization~\cite{zhou2022domain,Wang_2018}.
Domain adaptation focuses on adapting a model trained on source domain data to a target domain.
A common setting is unsupervised domain adaptation, where labeled source data is available, while target data is unlabeled.
Domain generalization addresses a harder setting, in which the model trains on labeled samples from several source domains, but is evaluated on previously unseen target domains. The goal is to learn a model that generalizes well to novel domains.

One central domain adaptation approach is discrepancy minimization, namely minimizing a distance measure between source and target distributions.
A prominent metric is Maximum Mean Discrepancy (MMD)~\cite{gretton2012kernel}, a kernel-based statistic for measuring the distance between two distributions by comparing moments.
Several works used MMD and its variants as domain adaptation objectives~\cite{long2015learning,long2017deep}, typically jointly with training a classifier on labeled source data.
One variant is local MMD (LMMD)~\cite{zhu2020deep}, which aligns class-specific subdomains rather than aligning full distributions. This requires using labels, but since the target domain is unlabeled, LMMD uses classifier predictions as pseudo-labels.

Although related, domain generalization includes several directions suited to the unseen target setting~\cite{zhou2022domain}.
One direction is domain alignment, which aims to minimize differences among multiple source domains to learn domain-invariant representations. 
This can be done using domain adaptation techniques, such as MMD-based objectives~\cite{li2018domain}.
We argue that the potential of these tools remains under-exploited in digital pathology.
In this work, we bridge this gap and show that fine-tuning PFMs with LMMD-based objectives 
can address both domain adaptation and generalization in digital pathology.

%% file: sec/3_method.tex
\section{Method}

\subsubsection{Domain Adaptation}

In unsupervised domain adaptation, 
$n_s$ labeled \emph{source} domain samples $\{(x_i^s,y_i^s)\}_{i=1}^{n_s}$ are given, 
where $x_i^s$ denotes a sample and $y_i^s$ is its label,
together with $n_t$ unlabeled \emph{target} domain samples $\{x_j^t\}_{j=1}^{n_t}$.
In our cross-hospital setting, samples are patch images, labels are biological classes (e.g., cancer subtypes), and each domain corresponds to a different hospital.
Our goal is to learn a patch classifier that achieves high accuracy on the target hospital despite having no access to its labels during training.
Note that source accuracy is not the objective, since source labels are available for supervised training.

To pursue this goal with a patch-level PFM $f_\theta$, we fine-tune the PFM parameters while learning a classifier $g_\phi$ from scratch.
The training objective has two components:
First, an LMMD loss~\cite{zhu2020deep} to align the source and target embedding distributions, designed to leverage the known domain mismatch between them.
Second, a supervised cross-entropy loss on the labeled source to preserve task-discriminative information (see~\cref{fig:da_method}).
Let $z_i^s = f_\theta(x_i^s)$ and $z_j^t = f_\theta(x_j^t)$ denote the source 
and target embeddings, respectively.
The training objective is
\begin{equation}
\label{eq:da_loss}
\begin{split}
\mathcal{L_{\mathrm{DA}}}(x^s, y^s, x^t)
&=
\frac{1}{n_s}\sum_{i=1}^{n_s}
\mathcal{L}_{\mathrm{CE}}\!\left(g_\phi(z_i^s),\, y_i^s\right) \\
&\quad + \lambda\, \mathcal{L}_{\mathrm{LMMD}}\!\left(\{(z_i^s, y_i^s)\}_{i=1}^{n_s},\ \{(z_j^t, g_\phi(z_j^t))\}_{j=1}^{n_t}\right),
\end{split}
\end{equation}
where $\mathcal{L}_{\mathrm{CE}}$ and $\mathcal{L}_{\mathrm{LMMD}}$ are the cross-entropy and LMMD losses, and $\lambda$ is a hyperparameter that controls the tradeoff between losses.
The LMMD term uses source labels $y_i^s$ and classifier predictions on target samples $g_\phi(z_j^t)$ to align class-specific subdomains.
Note that adaptation can be extended from a single source hospital to $K \geq 1$ hospitals by averaging per-source losses: $\frac{1}{K}\sum_{k=1}^{K}
\mathcal{L_{\mathrm{DA}}}\!\left(x^{s_k}, y^{s_k}, x^t\right)$.

\subsubsection{Domain Generalization}

In domain generalization, the setting is more challenging because the target domain is unknown during training. 
Here, samples from at least two source domains are given, and the goal is to learn a model that generalizes to previously unseen target domains.
To address this, we again use LMMD, but now apply it to align embedding distributions across pairs of source domains (see~\cref{fig:dg_method}).
The intuition is that encouraging similar representations across different source domains will reduce domain-specific variation and promote a representation that can transfer better to an unseen target domain. 
As the ground-truth labels of all sources are known, LMMD does not use the classifier predictions.
The training objective for $K \geq 2$ source domains is
\begin{equation}
\begin{split}
\mathcal{L_{\mathrm{DG}}}
&=
\frac{1}{K}\sum_{k=1}^{K}
\frac{1}{n_{s_k}}\sum_{i=1}^{n_{s_k}}
\mathcal{L}_{\mathrm{CE}}\!\left(g_\phi(z_i^{s_k}),\, y_i^{s_k}\right) \\
&\quad + \lambda\,
\frac{1}{|\mathcal{P}_K|}
\sum_{(a,b)\in\mathcal{P}_K}
\mathcal{L}_{\mathrm{LMMD}}\!\left(
\{(z_i^{s_a}, y_i^{s_a})\}_{i=1}^{n_{s_a}},
\{(z_j^{s_b}, y_j^{s_b})\}_{j=1}^{n_{s_b}}
\right),
\end{split}
\end{equation}
where $\mathcal{P}_K = \{(a,b)\,:\, 1 \le a < b \le K\}$ denotes all source pairs. Note that LMMD loss is symmetric in this case (i.e.\, it does not matter which source comes first).

\subsubsection{Slide-Level}

After fine-tuning the PFM with a patch-level objective we transfer it to slide-level classification by training a standard ABMIL 
classifier~\cite{ilse2018abmil} over embeddings produced by the adapted model. Training uses the source slides from which the source patches were extracted, along with their labels (see~\cref{fig:slide_method}).

%% file: sec/4_experiments.tex
\section{Experiments}

\subsection{Experimental Settings}
\subsubsection{Tasks and Datasets}

We evaluate PFM-LMMD on two subtyping tasks: non-small cell lung cancer (NSCLC; adenocarcinoma vs.\ squamous cell carcinoma) and renal cell carcinoma (RCC; clear cell vs.\ papillary). Patch-level data is derived from the TCGA~\cite{komura22tcga-ut} subset of the PathoROB dataset~\cite{komen2025robustfoundationmodelsdigital} (CC-BY-NC-SA 4.0). 
NSCLC and RCC comprise samples from 6 and 3 hospitals, respectively. 
For slide experiments, we use the TCGA slides from which patches are derived.

In domain adaptation, we focus on a setting with a single source hospital.
In domain generalization, we use 4 source hospitals.
Since the RCC data includes only 3 hospitals, we do not evaluate it under domain generalization.
At the patch level, for each task with $n$ hospitals, we sample a balanced set of 300 patches from each hospital. For each experiment, some sets are used as source domains and others as target domains. 
In domain adaptation, this results in 30 source–target combinations for NSCLC and 6 for RCC. Evaluation is performed on the unlabeled target set used for training. 
In domain generalization, we evaluate 15 NSCLC four-source combinations, each tested on the remaining unseen target hospitals.
For slide-level experiments, we train on source-hospital slides and evaluate on target-hospital slides. 
The number of slides per hospital varies (12–43),
since different patch-level sets originate from different numbers of slides.

\input{sec/tables_and_figures/table_main}

\subsubsection{Models and Training}

We evaluate PFM-LMMD on two PFMs: UNI~\cite{chen2024towards} and CONCH-v1.5~\cite{lu2024visual}.
We fine-tune PFMs using LoRA~\cite{hu2021lora} applied to the last half of the transformer blocks (rank 32), and jointly train a single-layer linear classifier. 
Training runs for 50 epochs. 
Each batch contains equal numbers of samples from each domain: 64 in domain adaptation, and 96 in domain generalization. 
A cosine learning rate~\cite{loshchilov2017sgdr} is used, starting from $10^{-3}$ for the classifier and $10^{-4}$ for LoRA layers. 
We use $\lambda=1.5$ in all experiments.
For slide-level experiments, we use standard tissue segmentation and patch extraction~\cite{zhang2025standardizing}, 
followed by training an ABMIL model~\cite{ilse2018abmil} for 30 epochs with a cosine learning rate starting from $10^{-4}$.

\subsubsection{Comparisons and Evaluation}
We compare PFM-LMMD to a linear classifier trained on original PFM embeddings using source-hospital data, as well as to Macenko~\cite{macenko2009method} and Reinhard~\cite{reinhard2002color} stain normalization (implementation by~\cite{barbano2022torchstain}), where models are trained on source images and evaluated on target images normalized to a source reference.
Our primary performance metric is balanced accuracy.

\subsection{Results}

\subsubsection{Patch-Level}

The patch-level domain adaptation results are shown in \cref{subtable:da_patch_level}.
PFM-LMMD improves over original PFMs in all settings, with gains from +3.24\% to +10.82\%.
Line 1 in~\cref{subtable:da_patch_level} reports in-domain results, where the classifier is trained and evaluated on samples from the same hospital. 
The drop from line 1 to 2 highlights the robustness problem.
PFM-LMMD mitigates it and in most cases surpasses the in-domain performance. 
In contrast, Reinhard and Macenko do not show a consistent effect, in line with previous findings~\cite{huang2025knowledge}.
PFM-LMMD improves the unadapted PFM in 93\% of patch-level domain adaptation source–target combinations across both tasks and PFMs. 
A one-sided Wilcoxon signed-rank test over all pairs confirms a statistically significant improvement ($p < 0.001$).
Similar gains are observed in macro F1 and AUROC. For example, on NSCLC with UNI, PFM-LMMD increases macro F1 by 6.52 points and AUROC by 3.57 points.

We further examine three practical domain adaptation scenarios. 
First, we evaluate generalization on a held-out target set not used during adaptation. Despite the increased difficulty, PFM-LMMD maintains performance gains (PFM-LMMD Gen.\  
in~\cref{subtable:da_patch_level}).
Second, while target data is unlabeled, our standard setup uses label-balanced source and target sets. 
To better reflect realistic conditions where the target distribution is unknown, we evaluate adaptation under a severe 70\%/30\% target label imbalance (PFM-LMMD Unbal.). Although gains are reduced, improvements remain consistent. 
Third, data from multiple source hospitals may be available, and we evaluate this setting for UNI on NSCLC (not in table): PFM-LMMD improves over UNI by +6.15\% on average.

Domain generalization results are reported in~\cref{subtable:dg_patch_level}. 
Gains are consistent, although smaller, as expected in this more challenging setting.

\subsubsection{Slide-Level}

\cref{subtable:da_slide_level} reports slide-level domain adaptation results.
PFM-LMMD outperforms the unadapted PFM, with gains of up to +11.96\%, although one case (RCC with CONCH-v1.5) shows limited improvement. 
These results are notable because patch-level fine-tuning does not include non-cancerous patches, whereas cancerous WSIs may contain substantial non-cancerous regions. 
The fact that PFM-LMMD obtains gains in this setting highlights its strong potential.

\subsection{Analysis and Ablations}

\input{sec/tables_and_figures/figure_analysis}

Beyond accuracy, PFM-LMMD demonstrates improved cross-hospital robustness in additional analyses.
\cref{fig:analysis} (left) shows a typical t-SNE visualization of original and adapted UNI embeddings on NSCLC, illustrating reduced hospital-driven clustering after adaptation. 
This effect is quantified using the inertia ratio between label-based and hospital-based clustering (lower is better): averaged over UNI NSCLC cases, PFM-LMMD reduces the ratio from 1.00 to 0.46.

We also evaluate robustness using the Robustness Index~\cite{komen2025robustfoundationmodelsdigital}, 
a patch-level metric that measures the capture of biological signals over hospital-related signals in $k$-nearest neighbors.
As shown in \cref{fig:analysis} (right), six representative adapted UNI models on NSCLC outperform the original UNI, even though the metric is computed across all hospitals in the data, not only those used for adaptation.

\cref{tab:analysis_small} compares LoRA fine-tuning with a projection head variant trained on fixed PFM embeddings. 
While the projection head improves over the original model, fine-tuning outperforms it in both domain adaptation and generalization.

%% file: sec/tables_and_figures/table_main.tex
\newcommand{\diff}[1]{{\fontsize{8}{9}\selectfont #1}}
\begin{table*}[!t]
    \centering
    \fontsize{8}{10.5}\selectfont
    \setlength{\tabcolsep}{3.8pt}

    \caption{
    Balanced accuracy (\%) results for
    (a) patch-level domain adaptation and (b) generalization, and (c) slide-level domain adaptation, averaged over experiments per setting. 
    Parentheses show differences from the original PFM.
    Gen.\ and Unbal.\ use different test sets; differences refer to the orig. PFM on these sets. 
    }
    \begin{subtable}{\linewidth}
    \centering
    \caption{Domain adaptation, patch-level.}
    \begin{tabular}{c|cc|cc}
        \toprule
        \multirow{2}{*}{\textbf{Method}} &
        \multicolumn{2}{c|}{\textbf{UNI}} &
        \multicolumn{2}{c}{\textbf{CONCH-v1.5}} \\
        & NSCLC & RCC &
        NSCLC & RCC \\
        \midrule
        \multicolumn{1}{l|}{Original PFM In-Dom.} & 
        82.72 & 96.77 &
        80.50 & 93.23 \\
        \midrule
        \multicolumn{1}{l|}{Original PFM} & 
        78.84 & 93.28 &
        76.15 & 86.03 \\
        \multicolumn{1}{l|}{Reinhard} & 
        78.23 \diff{(-0.61)} & 93.88 \diff{(+0.60)} &
        73.61 \diff{(-2.54)} & 86.72 \diff{(+0.69)} \\
        \multicolumn{1}{l|}{Macenko} & 
        76.16 \diff{(-2.68)} & 95.22 \diff{(+1.94)} &
        72.29 \diff{(-3.86)} & 90.32 \diff{(+4.29)} \\
        \multicolumn{1}{l|}{PFM-LMMD (ours)} & 
        \textbf{84.88} \diff{(+6.04)} & \textbf{96.52} \diff{(+3.24)} &
        \textbf{84.27} \diff{(+8.12)} & \textbf{96.85} \diff{(+10.82)} \\
        \midrule
        \multicolumn{1}{l|}{PFM-LMMD Gen.} & 
        84.33 \diff{(+5.33)} & 94.10 \diff{(+2.77)} &
        82.83 \diff{(+5.61)} & 93.65 \diff{(+9.82)} \\
        \midrule
        \multicolumn{1}{l|}{PFM-LMMD Unbal.} & 
        80.48 \diff{(+1.69)} & 94.40 \diff{(+1.93)} &
        79.00 \diff{(+2.53)} & 91.47 \diff{(+6.24)} \\
        \bottomrule
    \end{tabular}
    \label{subtable:da_patch_level}
    \end{subtable}
    \begin{subtable}{\linewidth}
    \centering
    \caption{Domain generalization, patch-level.}
    \begin{tabular}{c|c|c}
        \toprule
        \textbf{Method} &
        \textbf{UNI} (NSCLC) &
        \textbf{CONCH-v1.5} (NSCLC) \\
        \midrule
        \multicolumn{1}{l|}{Original PFM In-Dom.} & 
        82.96 &
        83.03 \\
        \midrule
        \multicolumn{1}{l|}{Original PFM} & 
        81.61 &
        80.45 \\
        \multicolumn{1}{l|}{Reinhard} & 
        79.82 \diff{(-1.79)} &
        80.01 \diff{(-0.44)} \\
        \multicolumn{1}{l|}{Macenko} & 
        78.84 \diff{(-2.77)} &
        76.66 \diff{(-3.79)} \\
        \multicolumn{1}{l|}{PFM-LMMD (ours)} & 
        \textbf{83.79} \diff{(+2.18)} &
        \textbf{82.46} \diff{(+2.01)} \\
        \bottomrule
    \end{tabular}
    \label{subtable:dg_patch_level}
    \end{subtable}
    \begin{subtable}{\linewidth}
    \centering
    \caption{Domain adaptation, slide-level.}
    \begin{tabular}{c|cc|cc}
        \toprule
        \multirow{2}{*}{\textbf{Method}} &
        \multicolumn{2}{c|}{\textbf{UNI}} &
        \multicolumn{2}{c}{\textbf{CONCH-v1.5}} \\
        & NSCLC & RCC &
        NSCLC & RCC \\
        \midrule
        \multicolumn{1}{l|}{Original PFM In-Dom.} & 
        81.48 & 89.19 &
        83.49 & 97.37 \\
        \midrule
        \multicolumn{1}{l|}{Original PFM} & 
        76.22 & 95.07 &
        82.29 & 93.85 \\
        \multicolumn{1}{l|}{Reinhard} & 
        72.91 \diff{(-3.31)} & 89.33 \diff{(-5.74)} &
        81.30 \diff{(-0.99)} & \textbf{94.48} \diff{(+0.63)} \\
        \multicolumn{1}{l|}{Macenko} & 
        71.88 \diff{(-4.34)} & 91.40 \diff{(-3.67)} &
        79.51 \diff{(-2.78)} & 93.85 \diff{(+0.00)} \\
        \multicolumn{1}{l|}{PFM-LMMD (ours)} & 
        \textbf{88.18} \diff{(+11.96)} & \textbf{97.17} \diff{(+2.10)} &
        \textbf{86.63} \diff{(+4.34)} & 92.93 \diff{(-0.92)} \\
        \bottomrule
    \end{tabular}
    \label{subtable:da_slide_level}
    \end{subtable}
    \label{table:patch_level}
\end{table*}

%% file: sec/tables_and_figures/figure_analysis.tex
\begin{figure}[t]
\centering
\begin{floatrow}
\floatsetup{valign=t,floatrowsep=columnsep}

\ffigbox[0.73\textwidth]{
\centering
\begin{subfigure}[t]{0.49\linewidth}
  \centering
  \raisebox{4mm}{\includegraphics[width=\linewidth]{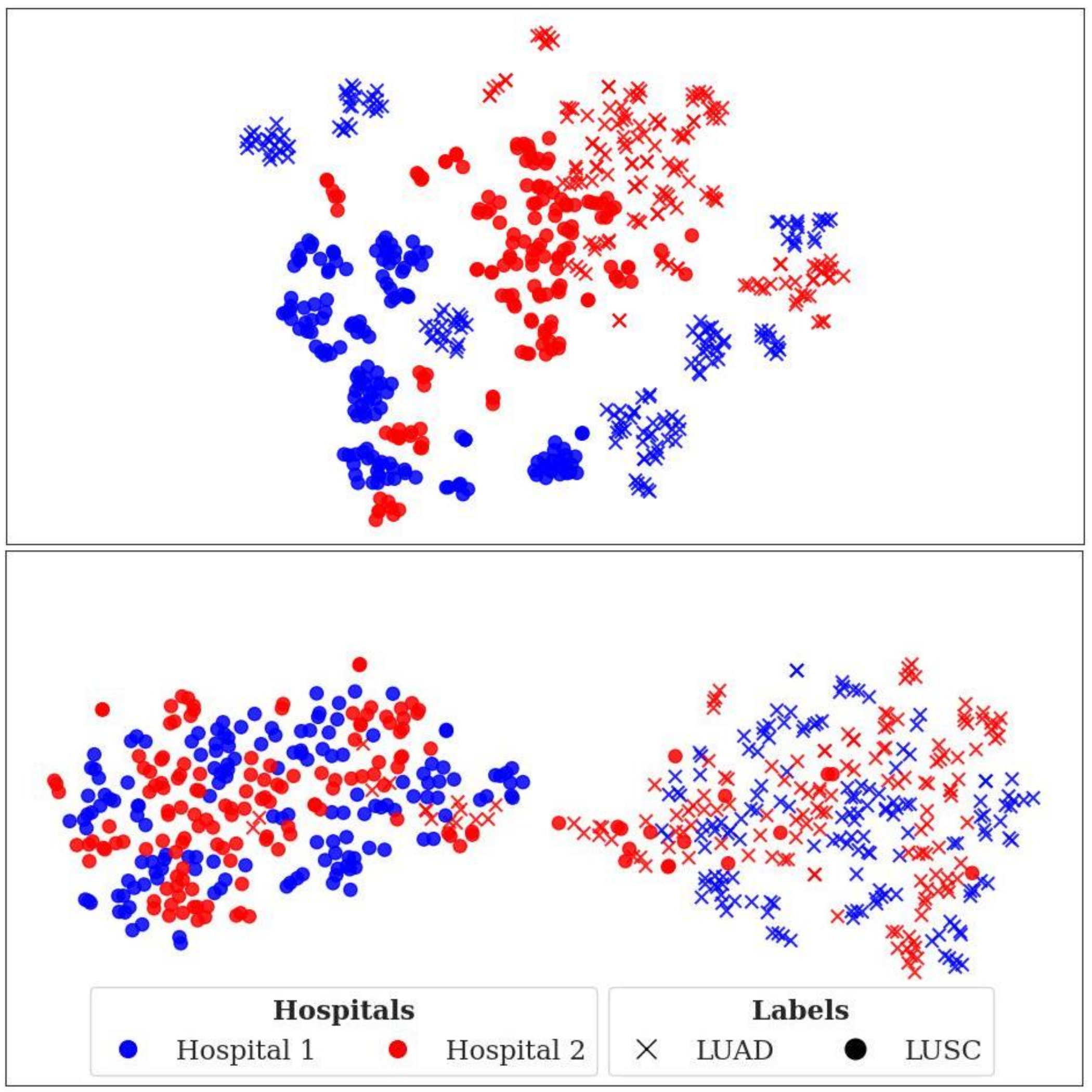}}
\end{subfigure}
\begin{subfigure}[t]{0.49\linewidth}
  \centering
  \includegraphics[width=\linewidth]{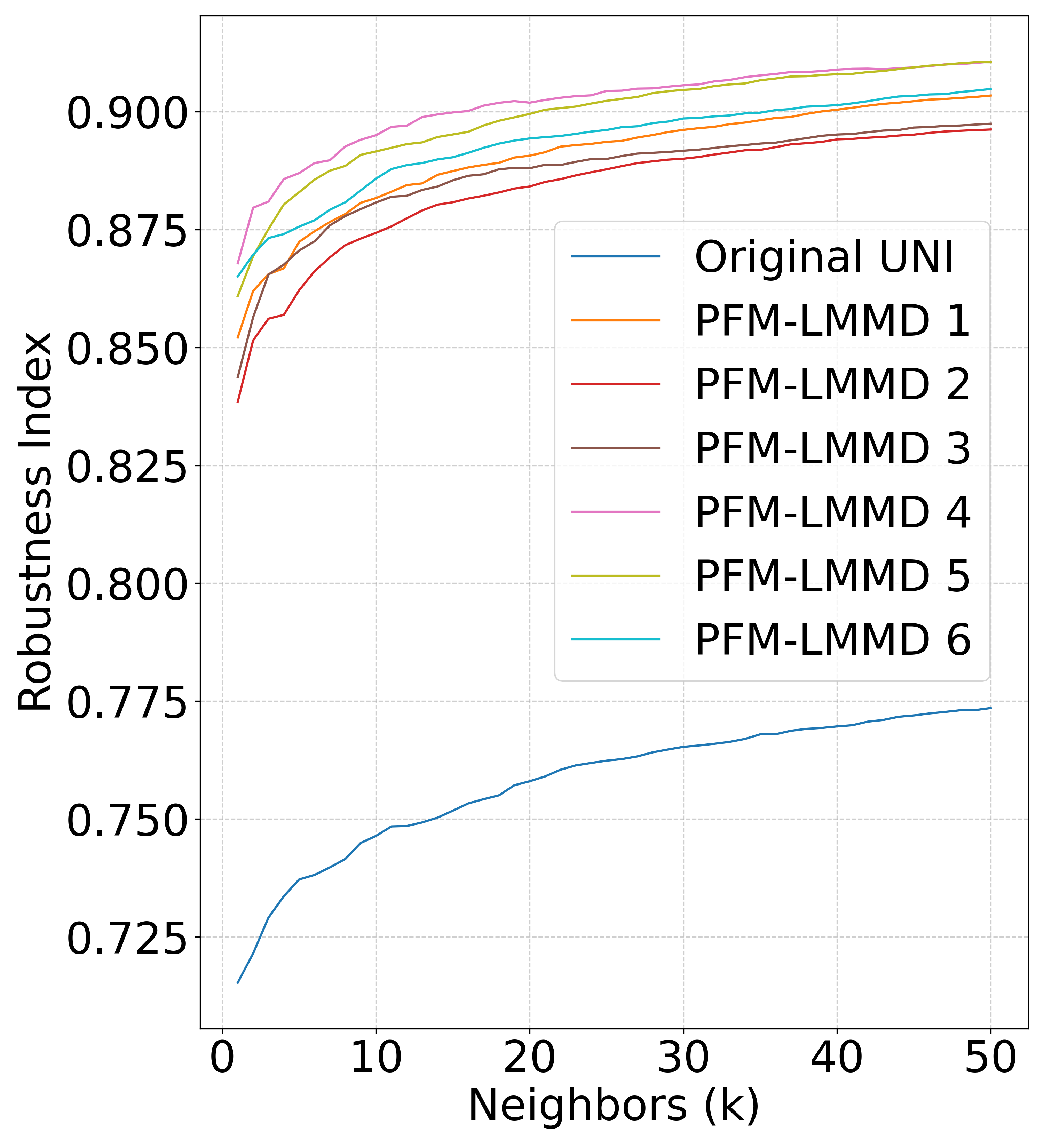}
\end{subfigure}
}{
\caption{Clustering of UNI (top-left) vs. adapted UNI (bottom-left), and Robustness Index comparison (right).
}
\label{fig:analysis}
}
\raisebox{35mm}{%
\ttabbox[0.23\textwidth]{
\centering
\fontsize{8}{10}\selectfont
\begin{tabular}{lcc}
\toprule
Method & DA & DG \\
\midrule
Original & 78.8 & 81.6 \\
No LoRA  & 83.9 & 82.5 \\
P-LMMD   & \textbf{84.9} & \textbf{83.8} \\
\bottomrule
\end{tabular}
}{
\caption{LoRA vs. learning on frozen embeddings.
}
\label{tab:analysis_small}
}}
\end{floatrow}
\end{figure}

%% file: sec/5_conclusions.tex
\section{Conclusion and Future Work}
\label{sec:conclusion}

We introduced PFM-LMMD, a lightweight framework that improves cross\hyp{}hospital robustness in PFMs through patch-level domain alignment using LMMD. 
The method is PFM-agnostic and applicable to both domain adaptation and domain generalization, with improvements transferring from patch-level training to the slide-level. Across experiments, PFM-LMMD achieves consistent gains.

Several directions remain for future work.
First, fine-tuning could explicitly incorporate healthy (non-cancerous) patches, which may further improve slide-level performance given the mixed composition of WSIs.
Second, while we focus on classification, similar cross-hospital robustness challenges likely arise in tasks such as regression, motivating extensions of PFM-LMMD to broader settings.